\documentclass[letterpaper, 10 pt, journal, twoside]{IEEEtran}  

\IEEEoverridecommandlockouts                              

\usepackage{amsmath,amsfonts}
\usepackage{amssymb}
\usepackage{amsthm}
\usepackage[ruled, lined, linesnumbered, commentsnumbered, longend]{algorithm2e}
\usepackage{array,multirow}
\usepackage{graphicx}
\usepackage[caption=false,font=small,labelfont=sf,textfont=sf]{subfig}
\usepackage[font=small,labelfont=bf]{caption}
\usepackage{textcomp}
\usepackage{stfloats}
\usepackage{url}
\usepackage{verbatim}
\usepackage{cite}
\usepackage{tabularx}
\usepackage{xcolor}
\usepackage{siunitx}
\usepackage{makecell}
\usepackage{comment}
\usepackage{float}
\usepackage{booktabs}
\usepackage{algpseudocode}
\usepackage{mathtools}
\usepackage{bm}
\usepackage{epsfig} 
\usepackage{times}
\usepackage{listings}
\usepackage{psfrag}
\usepackage[colorlinks,linkcolor=black,citecolor=black,urlcolor=black,bookmarks=false,hypertexnames=true]{hyperref} 
\usepackage{xparse}
\usepackage[english]{babel}
\usepackage{macros}
\usepackage{subfig}
\usepackage{graphicx}
\DeclareMathAlphabet\mathbfcal{OMS}{cmsy}{b}{n}

\title{\LARGE \bf
Good Deep Features to Track: \\Self-Supervised Feature Extraction and Tracking in Visual Odometry
}

\author{Sai Puneeth Gottam$^{1}$, Haoming Zhang$^{2}$,~\IEEEmembership{Member,~IEEE} and Eivydas Keras$^{2}$
\thanks{$^{1}$S. Gottam is with the Chair of Cyber-Physical-Systems, Montanuniversität Leoben, Leoben, Austria.}
\thanks{$^{2}$H. Zhang and E. Keras are with the Institute of Automatic Control, Faculty of Mechanical Engineering, RWTH Aachen University, Aachen, Germany.}
\thanks{\small Corresponding: haoming.zhang@rwth-aachen.de}
}

\begin{document}

\maketitle
\thispagestyle{empty}
\pagestyle{empty}

\begin{abstract}
Visual-based localization has made significant progress, yet its performance often drops in large-scale, outdoor, and long-term settings due to factors like lighting changes, dynamic scenes, and low-texture areas. These challenges degrade feature extraction and tracking, which are critical for accurate motion estimation. While learning-based methods such as SuperPoint and SuperGlue show improved feature coverage and robustness, they still face generalization issues with out-of-distribution data. We address this by enhancing deep feature extraction and tracking through self-supervised learning with task-specific feedback. Our method promotes stable and informative features, improving generalization and reliability in challenging environments.
\end{abstract}


\section{INTRODUCTION}
Visual-based robot localization has achieved remarkable success with a wide range of approaches tailored to specific applications. However, the performance of state-of-the-art approaches often degrades in large-scale and outdoor environments during a long-term operation, where a variety of challenges arise, such as dynamic lighting conditions and objects, low-parallax features and textureless surfaces. 

\begin{figure}[H]
\centering
\subfloat[Original image.]{\label{fig: demo_orin}\includegraphics[width=0.4\textwidth]{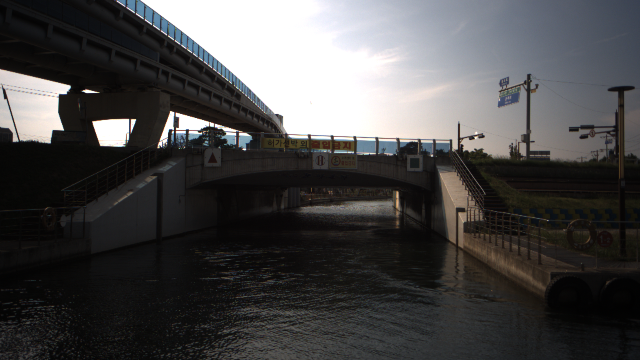}}
\hskip1ex\\ 
\subfloat[Feature extraction using SIFT \cite{sift}. ]{\label{fig: demo_sift}\includegraphics[width=0.4\textwidth]{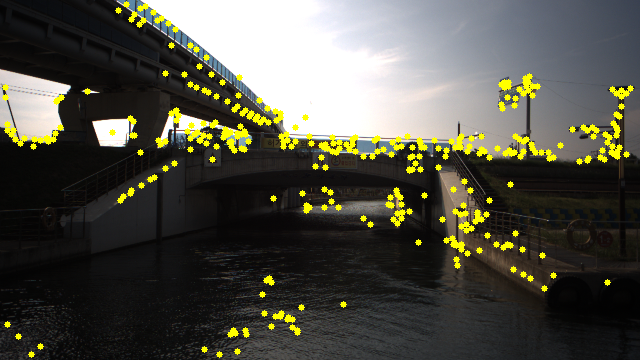}}
\hskip1ex\\ 
\subfloat[Feature extraction using GFTT \cite{gftt}. ]{\label{fig: demo_gftt}\includegraphics[width=0.4\textwidth]{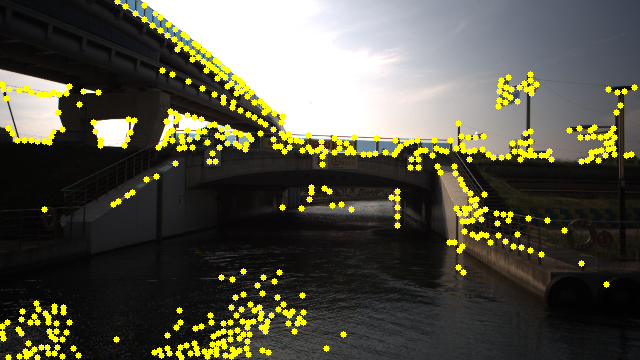}}
\hskip1ex\\ 
\subfloat[Feature extraction using pre-trained SuperPoint \cite{superpoint}. ]{\label{fig: demo_superpoint}\includegraphics[width=0.4\textwidth]{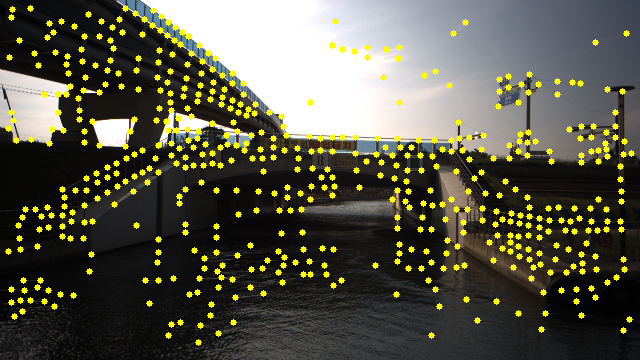}}
\caption{Extracted features (yellow) of visual odometry with different state-of-the-art approaches using the Pohang dataset (Pohang 04) \cite{pohang}.}
\label{fig: demo}
\end{figure}

These factors dramatically impact the quality and robustness of one of the most crucial steps in visual-based methods: feature extraction and tracking. Fig.\,\ref{fig: demo} exemplifies this issue in a maritime environment, where illumination and scene appearance can vary frequently even at the same location. 

Generally, a well-distributed coverage of tracked features across the entire image is expected to provide better geometric constraints. Assuming that the scale information is given, distant features help constrain rotation, while nearby features are essential for accurately resolving translational motion due to their stronger parallax \cite{soft2}. 

Keeping this in mind, classical feature extractors such as SIFT and GFTT tend to only focus on texture-rich and well-lit regions, resulting in densely clustered features and poor coverage across the image (see Fig.\,\ref{fig: demo_sift} and Fig.\,\ref{fig: demo_gftt}). 

In contrast, deep learning-based methods such as SuperPoint \cite{superpoint} produce more uniformly distributed results, even in low-texture or challenging lighting conditions (see Fig.\,\ref{fig: demo_superpoint}). However, since pre-trained model often degrade during inference when exposed to out-of-distribution data, obtaining high-quality features that can be robustly tracked to ensure accurate motion estimation in diverse real-world environments remains an open research question.

Although many methods exist to improve feature quantity — such as sophisticated feature matching and filtering \cite{soft2}, optimizing camera attributes (e.g., exposure time) \cite{bo_camera}, and removing dynamic objects \cite{dyn_vo} using semantic information \cite{dyn_slam} — they still struggle to generalize reliably for long-term motion estimation in diverse outdoor environments. 

Moreover, we observe that the distinction between “good” and “poor” visual features is not always clear-cut, as it depends not only on geometric properties but also on factors such as scale and temporal consistency. In visually degraded scenarios, features previously considered “poor” may still be trackable over short time windows and, in some cases, are sufficient to prevent visual odometry (VO) failure.

Given these observations, we aim to improve the performance and generalization of the deep learning-based feature extractor and tracker by leveraging self-supervised learning with well-designed loss terms that provide situational feedback on what constitutes “good deep features to track”. We chose SuperPoint \cite{superpoint} and SuperGlue \cite{superglue} as our backbone networks compared to other discriminative models, as they are capable of capturing structurally and semantically meaningful features thanks to their adaptive (e.g., homographic adaptation) training strategy.  

Unlike \cite{RL_VO}, which treats the VO as a black box and only tunes hyperparameters (e.g., grid size) from a system-level perspective, we argue that it is (more) important to focus on the core components of a VO. 

We build on these foundations by directly modifying and fine-tuning the core components of feature extraction and matching in a self-supervised way. Our work integrates recent datasets like Pohang \cite{pohang} and builds upon proven frameworks such as RTAB-Map \cite{rtabmap}.

Our hypothesis is supported by preliminary results, which show that the proposed method can effectively update the model parameters and force the models to ignore unstable or uninformative features. 
\begin{figure*}[!t]
    \centering
    \includegraphics[width=0.9\textwidth]{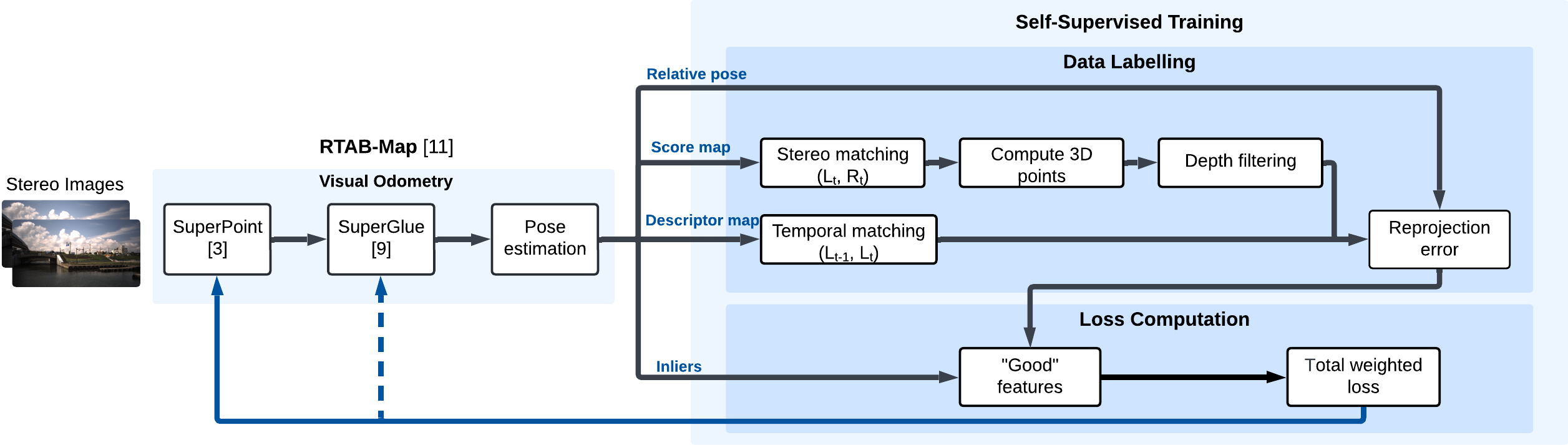}
    \caption{Proposed self-supervised pipeline to train the SuperPoint \cite{superpoint} and SuperGlue \cite{superglue} models within a well-established visual SLAM framework RTAB-Map \cite{rtabmap}.}
    \label{fig: flowchart}
\end{figure*}

\section{Method}
We build on the RTAB-Map framework by replacing the feature extractor and matcher with SuperPoint \cite{superpoint} and SuperGlue \cite{superglue}. 

Fig.\,\ref{fig: flowchart} illustrates the proposed self-supervised learning pipeline in our work. We employed RTAB-Map \cite{rtabmap} as the baseline visual odometry, where the feature extractor and tracker are replaced by SuperPoint \cite{superpoint} and SuperGlue \cite{superglue}. 

During training, estimated relative pose, stereo and temporal correspondences are used to evaluate reprojection errors, which serve as geometric consistency cues. Features that are well-tracked and geometrically reliable are identified as “good” and used to compute a total weighted loss:
\begin{align}
\begin{split}
    L &= \underbrace{w_i \cdot L_i(\myFrameVec{X}{}{}, \myFrameVec{Y}{}{}) + w_i^{\prime} \cdot L_i(\myFrameVec{X}{}{\prime}, \myFrameVec{Y}{}{\prime}) + w_{\rm pk} \cdot L_{\rm pk}}_{\text{training detector head}} +\\ 
    &+\underbrace{w_{d} \cdot L_{d}(\mathbfcal{D}, \mathbfcal{D}^{\prime}, \mathbfcal{S})}_{\text{training descriptor head}},
\end{split}
\end{align}
where the loss terms $L_i(\myFrameVec{X}{}{}, \myFrameVec{Y}{}{})$ and $L_i(\myFrameVec{X}{}{\prime}, \myFrameVec{Y}{}{\prime})$ penalize non-keypoint score map $\myFrameVec{X}{}{}$ with their labels $\myFrameVec{Y}{}{}$ from image $[\myFrameVec{X}{}{},~\myFrameVec{Y}{}{}]\in\myFrameVec{I}{}{}$, and the corresponding warped image $[\myFrameVec{X}{}{\prime},~\myFrameVec{Y}{}{\prime}]\in\myFrameVec{I}{}{\prime}$. We use a peaky loss $L_{\rm pk}$ to penalize inconsistent or diffuse feature responses in the score maps, which are often caused by blur or ambiguous textures. In addition, a hinge loss $L_{d}$ is employed to enforce descriptor discriminability by maximizing the distance between non-matching features and minimizing it for matching pairs.

\section{Preliminary Results and Discussion}
\subsection{Feature Extraction}
\begin{figure*}[!h]
    \centering
    \includegraphics[width=1\textwidth]{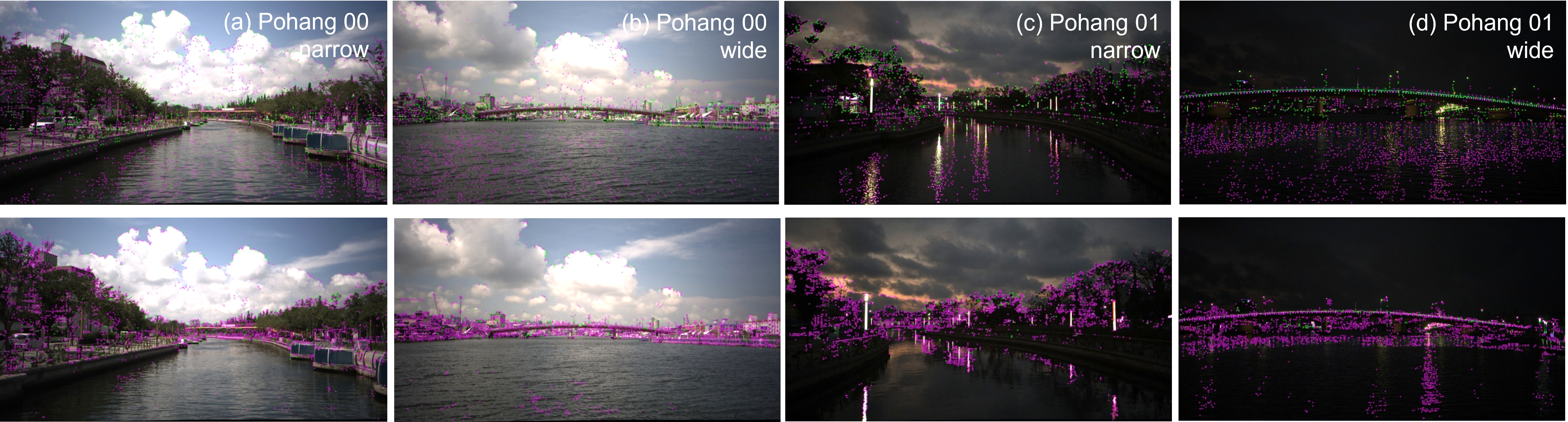}
    \caption{Extracted visual features (purple and green) of Pohang data \cite{pohang} in different scenes and datetime using original pre-trained (top row) and self-supervised (bottom row) SuperPoint model. }
    \label{fig: res_pohang}
\end{figure*}

\begin{figure}[!h]
\centering
\subfloat[Features from pre-trained SuperPoint.]{\label{fig: res_kitti_pre}\includegraphics[width=0.48\textwidth]{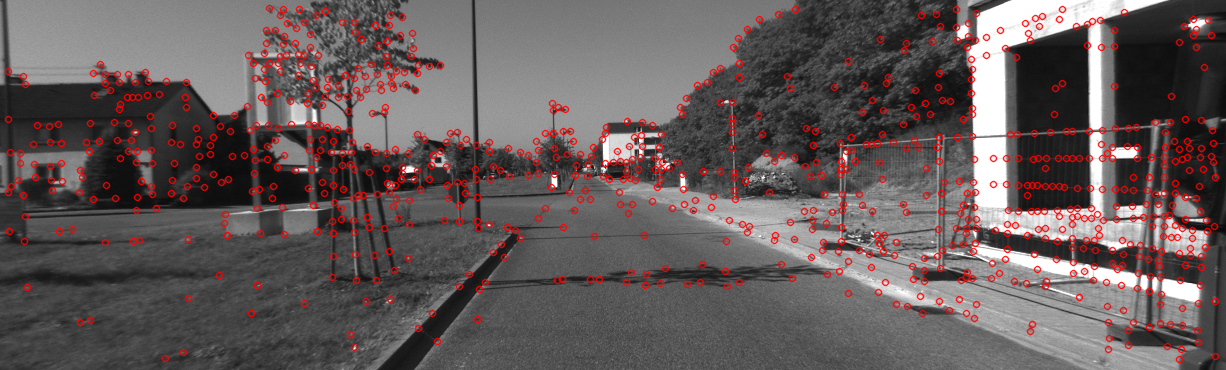}}
\hskip1ex\\ 
\subfloat[Features from self-supervised SuperPoint.]{\label{fig: res_kitti_post}\includegraphics[width=0.48\textwidth]{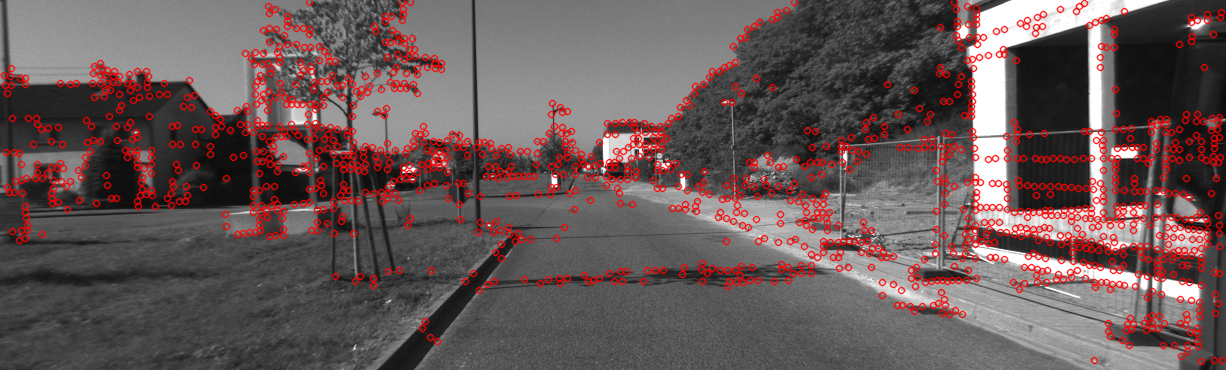}}
\caption{Extracted visual features (red) of KITTI data (seq. 06) \cite{kitti}.}
\label{fig: res_kitti}
\end{figure}

Fig.\,\ref{fig: res_kitti} and Fig.\,\ref{fig: res_pohang} present qualitative results of feature extraction using the pre-trained SuperPoint model and the proposed self-supervised version. 

Compared to the former, features extracted by the self-supervised model tend to be more equitably spread across expected regions in the image, such as rigid structures, while suppressing features from less reliable areas like terrain and water surfaces.

\subsection{Trajectory Estimation}
To validate the proposed method, we use both the pre-trained and self-supervised SuperPoint models as feature extractors within the RTAB-Map framework to estimate the vessel’s trajectory. In addition, we employ the trajectory estimation using ORB features \cite{orb} as a baseline method for comparison. 

We evaluate all methods on the Pohang dataset \cite{pohang}, which is divided into sequences recorded in narrow and wide water canals, during both daytime and nighttime (see Fig. 4). The results are presented in Fig.5.

\begin{figure}[!h]
\centering
\subfloat[Trajectories in narrow canal during daytime.]{\label{fig: res_traj_narrow_daytime}\includegraphics[width=0.42\textwidth]{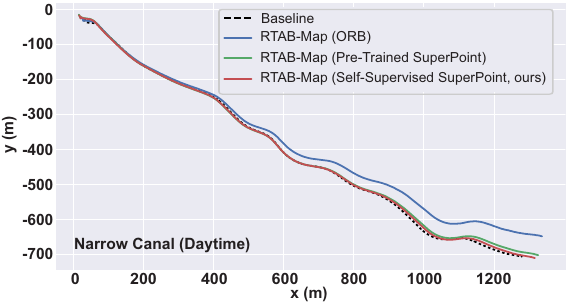}}
\hskip1ex\\ 
\subfloat[Trajectories in narrow canal during nighttime.]{\label{fig: res_traj_narrow_nighttime}\includegraphics[width=0.44\textwidth]{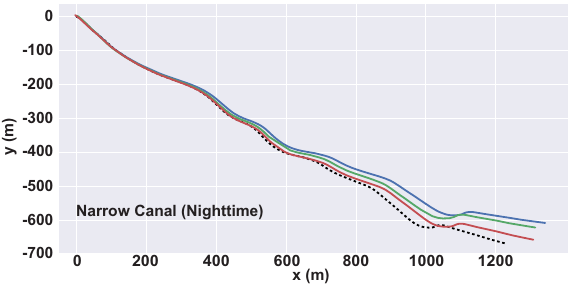}}
\hskip1ex\\ 
\subfloat[Trajectories in off-shore area during daytime.]{\label{fig: res_traj_wide_daytime}\includegraphics[width=0.44\textwidth]{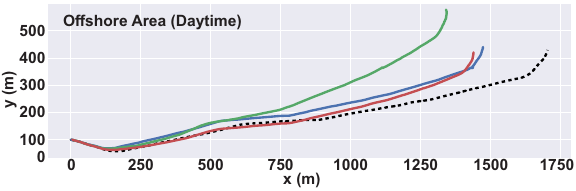}}
\hskip1ex\\ 
\subfloat[Trajectories in off-shore area during nighttime.]{\label{fig: res_traj_wide_nighttime}\includegraphics[width=0.44\textwidth]{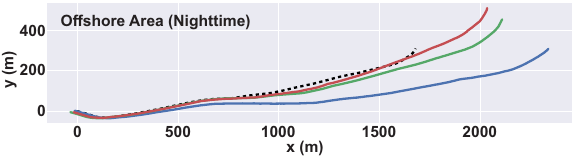}}
\caption{Estimated trajectories of both narrow canal and offshore areas in Pohang \cite{pohang} using different feature extractors implemented in RTAB-Map \cite{rtabmap}.}
\label{fig: res_trajectory}
\end{figure}

In narrow areas (Fig.4a and 4c) where most features can be acquired from infrastructure objects during daytime, the proposed model outperforms the others. A similar trend is observed during nighttime, although all VO methods experience degradation due to insufficient visual features. 

The problem of insufficient visual features becomes more pronounced in near-offshore areas. Surprisingly, the estimated trajectories using SuperPoint features exhibit less drift during nighttime compared to daytime, which stands in contrast to the behavior observed with ORB features. This may be due to fewer visual distractions at night, such as reflections and moving elements, which are more common during the day. In such cases, classic feature extractors, which are unable to uniformly explore the entire image, may occasionally avoid distracting regions. 
\clearpage
\section{Conclusion and Future Work}
This work aims to enhance the generalization of deep learning-based feature extraction and tracking within a self-supervised training framework. In the early stage of this research, we successfully demonstrate a proof of concept, showing that the model can be effectively improved using simple loss terms—without relying on label-intensive training procedures.

However, we did not train the feature matcher (e.g., SuperGlue), which remains pre-trained on the original SuperPoint. Our current results are also \textbf{not yet sufficient to fully answer the question of what defines “good deep features” to track}. 

\begin{figure}[!h]
\centering
\subfloat[\texttt{gnssFGO} without visual odometry.]{\includegraphics[width=0.48\textwidth]{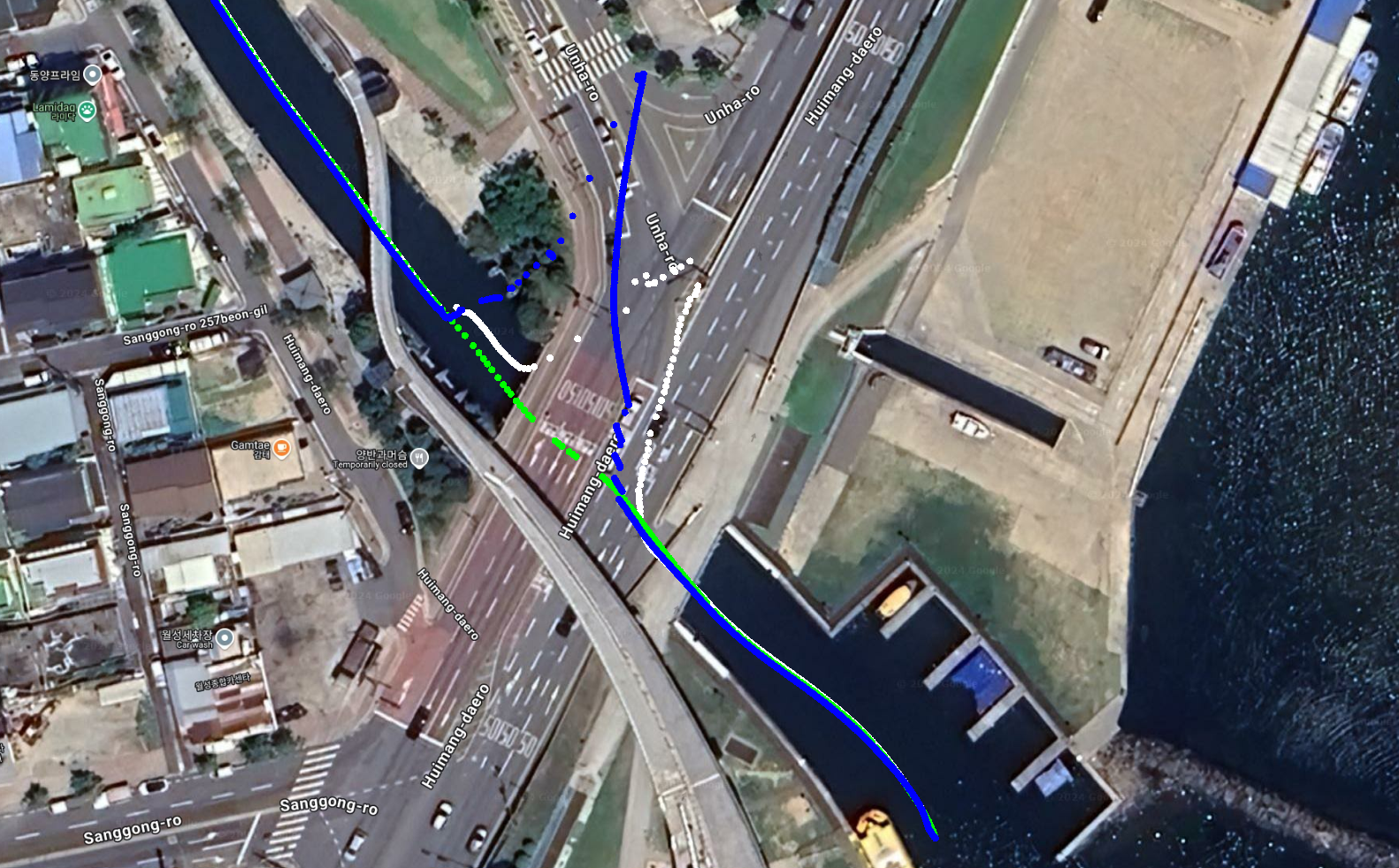}}
\hskip1ex\\ 
\subfloat[\texttt{gnssFGO} with visual odometry.]{\includegraphics[width=0.48\textwidth]{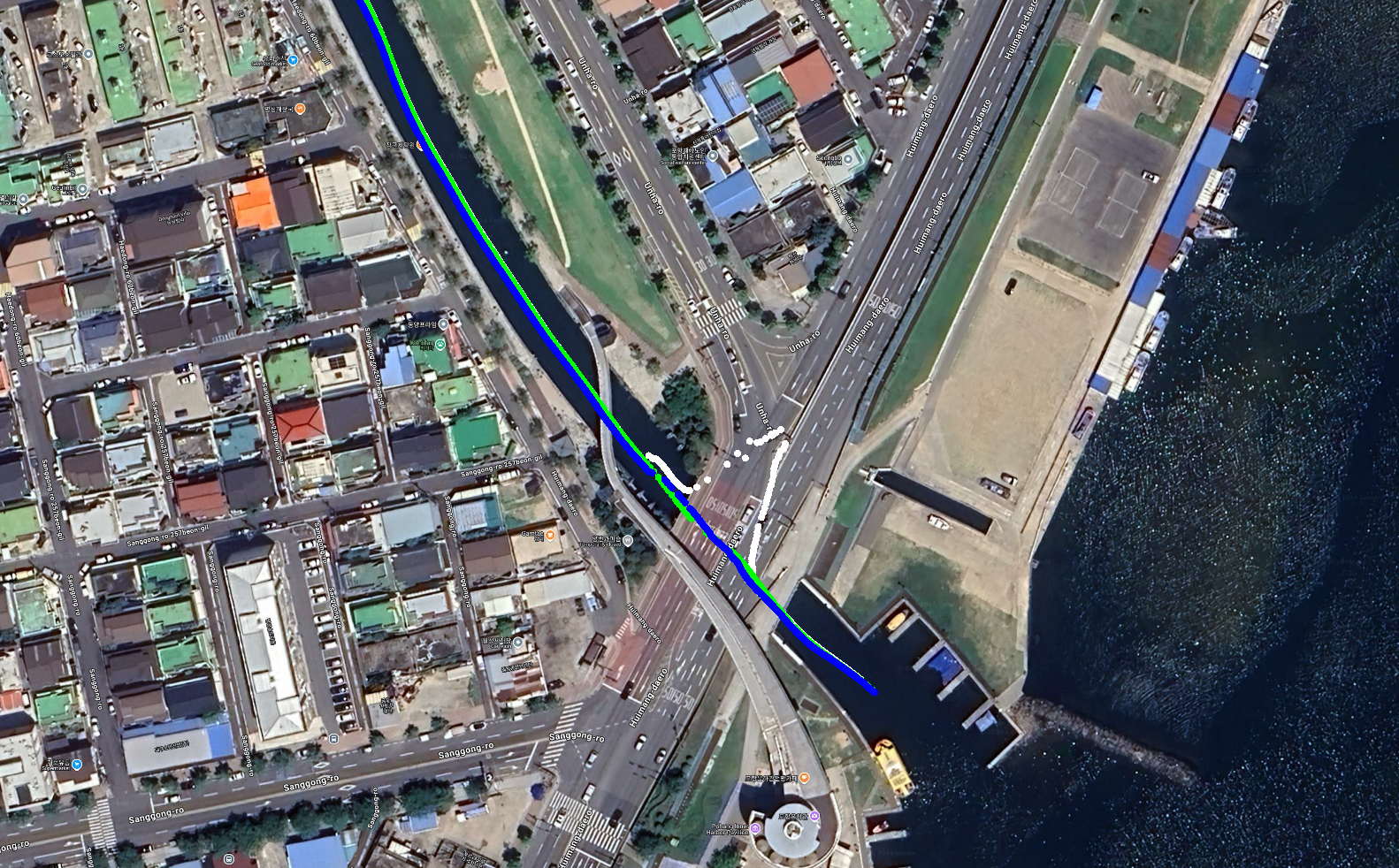}}
\caption{Demonstration of robot localization using \texttt{gnssFGO} \cite{gnssfgo}. The green, white, and blue trajectories represent the baseline, GNSS standalone, and \texttt{gnssFGO} estimates.}
\label{fig: res_gnssfgo}
\end{figure}

Additionally, since vision-based localization faces more challenges in outdoor environments than modalities like GNSS or LiDAR, we aim to explore \textbf{how VO can be better designed and integrated into a multisensory state estimator for resilient, long-term robot navigation} (see Fig.\,\ref{fig: res_gnssfgo}).

\vfill

\section*{Acknowledgment}
The presented research was supported by the German Federal Ministry of Economic Affairs and Climate Action (BMWK) under Project 19F1150B (Museas). 
The work was conducted at the Institute of Automatic Control (IRT), RWTH Aachen University. The authors also thank Space Applications Services NV/SA in Belgium for their support for providing hardware support during this work.

\addtolength{\textheight}{-12cm}   






\bibliographystyle{IEEETran.bst}
\bibliography{reference}

\end{document}